\documentclass[10pt,twocolumn,letterpaper]{article}

 \usepackage{cvpr}              
\usepackage{booktabs}
\usepackage{graphicx}
\usepackage{amsmath}
\usepackage{amssymb}
\usepackage{xcolor}
\usepackage[utf8]{inputenc}
\usepackage{textgreek}
\usepackage{booktabs}
\usepackage{multirow}
\usepackage{graphicx}
\usepackage{tabularx}
\usepackage{makecell}
\usepackage{placeins}
\usepackage{balance} 
\usepackage{textalpha}
\definecolor{cvprblue}{rgb}{0.21,0.49,0.74}
\usepackage[pagebackref,breaklinks,colorlinks,allcolors=cvprblue]{hyperref}

\def\paperID{*****} 
\def\confName{CVPR}
\def\confYear{2026}

\title{YOLO26 vs. YOLOv8: A Comprehensive Architectural Benchmark of Next-Generation Real-Time Object Detection Models}

\author{
Chidera G. Oguine$^{1}$,
Kanyifeechukwu J. Oguine$^{2}$,
Obiozor M. Oguine$^{1}$\\
Ozioma C. Oguine$^{3}$\\
$^{1}$University of Abuja, Nigeria
\quad
$^{2}$Vanderbilt University, USA
\quad
$^{3}$University of Notre Dame, USA\\
\texttt{cg.oguine@uniabuja.edu.ng}
}

\begin{document}
\maketitle
\begin{abstract}
This paper presents a rigorous empirical evaluation of Ultralytics YOLO26 against the YOLOv8 baseline, offering an independent real-world stress test of NMS-free architectures on non-COCO distributions. Engineered for edge deployment, YOLO26 introduces native end-to-end one-to-one label assignment, the removal of Distribution Focal Loss (DFL), and a spectral-constrained CSP-Muon backbone. We execute a comprehensive, cross-scale comparative analysis across five model capacities utilizing general object detection (Pascal VOC) and dense aerial small-object detection (VisDrone) datasets. Models are evaluated across accuracy ($mAP_{50}$ and $mAP_{50:95}$), model complexity, and hardware-specific CPU/GPU latency. Our findings reveal that while YOLO26 achieves a lower computational footprint and superior accuracy on Pascal VOC with YOLO26-x reaching 0.635 $mAP_{50:95}$. This advantage narrows in dense aerial environments. On VisDrone, where over 75\% of objects are under 2,000 pixels, both architectures struggle significantly, yielding a minimal performance gap (0.214 $mAP_{50:95}$ for YOLOv8-x vs. 0.224 $mAP_{50:95}$ for YOLO26-x). Crucially, hardware benchmarking demonstrates that YOLOv8 maintains a consistent edge in GPU inference latency across identical scales (e.g., 6.92 ms for YOLOv8-s vs. 8.38 ms for YOLO26-s), proving that NMS-free design does not inherently guarantee universal deployment superiority. Ultimately, this work maps the operational boundaries of NMS-free frameworks to guide architecture selection based on dataset density, object scale, and hardware constraints.

\end{abstract}

\section{Introduction}
\label{sec:intro}

Real-time object detection underpins critical applications spanning autonomous navigation \cite{janai2020computer}, aerial surveillance \cite{zhan2022improved}, and industrial quality control \cite{ahmed2024real}. Since its inception in 2016, the You Only Look Once (YOLO) framework \cite{redmon2016you} has defined the state-of-the-art for single-shot detection, evolving through successive iterations that progressively refined speed-accuracy trade-offs. However, a persistent limitation has constrained every generation from YOLOv1 \cite{redmon2016you}  through YOLO11 \cite{yolo11_ultralytics}, the reliance on Non-Maximum Suppression (NMS) as a post-processing heuristic to filter redundant bounding box predictions. NMS operates by iteratively selecting the highest-confidence detection and suppressing overlapping alternatives based on a fixed Intersection-over-Union (IoU) threshold \cite{hosang2017nms}. While effective at deduplication, this process introduces three critical deficiencies: (1) latency variance proportional to scene density \cite{wang2024yolov10}, (2) hyperparameter sensitivity requiring per-deployment tuning of the IoU threshold, and (3) hardware incompatibility on integer-arithmetic accelerators where sequential filtering disrupts parallel execution graphs \cite{lv2024rt}.

YOLO26 \cite{yolo26_ultralytics}, introduced by Ultralytics in early 2026, fundamentally resolves these limitations through a native end-to-end architecture that eliminates NMS entirely. Building on the dual-label assignment concepts pioneered by YOLOv10 \cite{wang2024yolov10}, YOLO26 redesigns the detection head to enforce one-to-one label assignment during training, producing deterministic, non-redundant predictions without post-processing. Complementing this architectural shift are three training innovations: the MuSGD optimizer for stabilizing lightweight backbones, Small-Target-Aware Label Assignment (STAL) for improving detection of objects occupying less than 1\% of image area, and Progressive Loss Balancing (ProgLoss) for dynamic supervision scheduling \cite{jordan2024muon, yolo26_ultralytics}.

Despite these theoretical advantages, the empirical performance of YOLO26 relative to its high-impact predecessor YOLOv8 \cite{yolov8_ultralytics} remains insufficiently characterized on datasets that stress specific failure modes of real-world deployment. The MS COCO benchmark \cite{lin2014microsoft}, while standard, emphasizes medium-to-large objects in everyday scenes; it does not adequately test the small-object fidelity critical to aerial surveillance or the class-generalization demands of legacy taxonomies. This study addresses this gap by evaluating both models on Pascal VOC \cite{everingham2010pascal}, a foundational benchmark for general object detection, and VisDrone \cite{zhu2021visdrone}, a UAV-captured dataset where over 75\% of objects occupy fewer than 0.1\% of image pixels.

The contributions of this paper are threefold: \textbf{(1)} a comprehensive architectural and mathematical comparison of YOLOv8 and YOLO26, including their loss functions and post-processing pipelines; \textbf{(2)} empirical benchmarking on Pascal VOC and VisDrone with detailed performance metrics; and \textbf{(3)} an analysis of deployment-context suitability that challenges the assumption that newer architectures universally dominate their predecessors across all hardware and data distributions.
\section{Related Work}
\label{sec:related_work}

\subsection{Real-Time Object Detectors and the Evolution of YOLO}
The YOLO (You Only Look Once) family has undergone continuous architectural refinement since \citet{redmon2016you} introduced the original single-shot formulation in 2016. \autoref{tab:yolo-evolution} summarizes the evolutionary trajectory through the 2026 flagship, YOLO26. The lineage is defined by a transition from anchor-based to anchor-free designs, the introduction of decoupled heads, and the recent pivot toward NMS-free end-to-end inference.

\begin{table*}[t]
\centering
\caption{Architectural Evolution of the YOLO Family (v1-v26)}
\label{tab:yolo-evolution}
\resizebox{\textwidth}{!}{
\begin{tabular}{@{}cllllllll@{}}
\toprule
\textbf{Model} & \textbf{Year} & \textbf{Backbone} & \textbf{Neck} & \textbf{Head} & \textbf{Anchors} & \textbf{Loss Components} & \textbf{Post-Proc.} & \textbf{Key Innovation} \\
\midrule
YOLOv1 \cite{redmon2016you} & 2016 & Darknet-19 & --- & Single & Yes & MSE + CE & NMS & First unified detection network \\
YOLOv2 \cite{redmon2017yolo9000} & 2017 & Darknet-19 & --- & Single & Yes & MSE + CE & NMS & Anchor boxes, batch normalization \\
YOLOv3 \cite{redmon2018yolov3} & 2018 & Darknet-53 & FPN & Multi-scale & Yes & BCE + MSE & NMS & Feature pyramid, multi-label \\
YOLOv4 \cite{bochkovskiy2020yolov4} & 2020 & CSPDarknet53 & PANet & Multi-scale & Yes & CIoU + BCE & NMS & Bag of freebies, mosaic aug. \\
YOLOv5 \cite{yolov5} & 2020 & CSPDarknet & PANet & Multi-scale & Yes & BCE + CIoU & NMS & Modular PyTorch, auto-anchor \\
YOLOv8 \cite{yolov8_ultralytics} & 2023 & C2f & PANet & Decoupled & No & BCE + CIoU + DFL & NMS & Anchor-free, decoupled head \\
YOLOv9 \cite{wang2024yolov9} & 2024 & GELAN & PANet & Decoupled & No & BCE + CIoU + DFL & NMS & Programmable Gradient Info. \\
YOLOv10 \cite{wang2024yolov10}& 2024 & GELAN & PANet & Decoupled & No & BCE + CIoU + DFL & NMS-Free & Dual-label assignment, PSA \\
YOLO11 \cite{yolo11_ultralytics} & 2024 & C3k2 & PANet & Decoupled & No & BCE + CIoU + DFL & NMS & C2PSA attention, multi-task \\
YOLO12 \cite{tian2025yolo12} & 2025 & Flash + A2A$^2$ & PANet & Decoupled & No & BCE + CIoU + DFL & NMS & Area Attention, R-ELAN \\
YOLO13 \cite{yolov13} & 2025 & Hyper-Net & PANet & Decoupled & No & BCE + CIoU + DFL & NMS & Hypergraph spatial modeling \\
\midrule
\textbf{YOLO26} \cite{yolo26_ultralytics} & \textbf{2026} & \textbf{CSP-Muon} & \textbf{PANet} & \textbf{Decoupled (1-to-1)} & \textbf{No} & \textbf{STAL + ProgLoss} & \textbf{NMS-Free} & \textbf{Edge-optimized, DFL-free} \\
\bottomrule
\end{tabular}
}
\end{table*}

Recent developments in the YOLO family reveal a clear architectural divergence beginning around 2024. While YOLOv10 introduced end-to-end, NMS-free training through consistent dual-label assignment, subsequent iterations such as YOLO11 \cite{yolo11_ultralytics} - YOLO13 \cite{yolov13} retained Non-Maximum Suppression (NMS) during inference to preserve compatibility with existing deployment pipelines and tooling. In contrast, YOLO26 fully adopts an end-to-end inference paradigm by eliminating both NMS and Distribution Focal Loss (DFL). The removal of DFL is particularly significant for edge deployment because DFL introduces additional computational overhead and quantization complexity, which can negatively affect inference efficiency on resource-constrained hardware.

Prior studies \cite{sapkota2025yolo26, oguine2022yolo, yolov8_ultralytics, bochkovskiy2020yolov4} have extensively benchmarked YOLO variants on the MS COCO dataset. For example, \citet{sapkota2025yolo26} evaluated YOLOv5 through YOLO26 and demonstrated that YOLO26 achieves improved speed-accuracy trade-offs under TensorRT-based GPU inference. Furthermore, a foundational study by \citet{terven2023comprehensive} established that while architectural scaling (increasing depth and width) provides diminishing returns for mAP, the optimization of the ``neck" and ``head" components specifically through the removal of redundant post-processing offers the most significant gains in real-world throughput. These studies collectively established standardized benchmarking practices centered on metrics such as mAP, latency, parameter efficiency, and inference throughput, thereby providing a foundation for evaluating architectural changes across YOLO generations. However, existing evaluations primarily focus on server-grade GPU environments and may not generalize to CPU-bound edge systems or datasets characterized by severe object-scale imbalance. Prior independent evaluations of YOLOv8 on VisDrone2019-DET  by \citet{tan2025dbyolov8} and \citet{hu2025mff} reported mAP@50 ranging from 37.3\% (YOLOv8s) to 44.3\% (YOLOv8x) under standard pretrained training configurations, with mAP@50:95 between 22.1\% and 27.2\%. These results underscore the persistent challenge of small-object detection in UAV imagery, where even state-of-the-art general-purpose detectors struggle to exceed 30\% mAP@50:95.

\subsection{Pascal VOC and VisDrone Datasets}
The Pascal Visual Object Classes (VOC) challenge, established in 2005 and concluding in 2012, provides a foundational benchmark for object detection evaluation \cite{everingham2010pascal}. The combined VOC 2007/2012 dataset contains 16,551 training images and 4,952 test images across 20 object categories. Unlike the 80 classes found in MS COCO \cite{lin2014microsoft}, VOC's taxonomy emphasizes everyday objects with relatively large, well-centered instances. While PASCAL-VOC remains a foundational benchmark, recent work by \citet{tong2023rethinking} demonstrates that persistent labeling errors, including missing annotations in occluded scenes and category mislabeling, significantly degrade small-object detection performance. Through visual analysis, they identify median relative object areas ranging from 1.38\% to 46.4\%, achieving AP below 80\% even on modern architectures. These findings suggest that VOC's apparent ``solved'' status masks underlying data quality challenges that become pronounced when evaluating detectors on small or occluded instances.

The VisDrone2019-DET dataset comprises 10,209 static images captured by UAVs across 14 cities under diverse environmental conditions \cite{zhu2021visdrone}. It annotates 10 categories, including pedestrians, vehicles, and various tricycle types. Approximately 75\% of objects occupy fewer than 2,000 pixels, and 97\% account for less than 1\% of image area, presenting a significant ``small-object" challenge \cite{zhu2021visdrone}. Recent evaluations on UAV datasets by \citet{zhan2022improved} highlight the challenges of detecting densely packed small objects, where occlusion and 
spatial congestion cause missed detections and false positives. In such high-density scenarios, the proximity of candidate bounding boxes complicates IoU-based suppression heuristics, frequently triggering excessive suppression of distinct, valid adjacent targets. This failure mode motivates the adoption 
of NMS-free architectures \cite{wang2024yolov10} for deterministic inference in congested aerial environments. \autoref{tab:dataset-comparison} compares these datasets across key dimensions relevant to detector evaluation.

\begin{table*}[t]
\centering
\caption{Dataset Characteristics: Pascal VOC vs. VisDrone}
\label{tab:dataset-comparison}
\begin{tabular}{@{}lcc@{}}
\toprule
\textbf{Characteristic} & \textbf{Pascal VOC 2007+2012} & \textbf{VisDrone2019-DET} \\
\midrule
Total Images & 21,503 & 10,209 \\
Train / Val / Test & 16,551 / --- / 4,952 & 6,471 / 548 / 1,610 \\
Resolution & $\sim$500$\times$375 (Variable) & Up to 2,000$\times$1,500 (Variable) \\
Object Categories & 20 & 10 \\
Avg. Objects/Image & 2.3 & 34.6 \\
Small Objects ($<$32$\times$32 px) & $<$5\% & $>$75\% \\
Scene Type & General photography & Aerial/UAV surveillance \\
Occlusion Frequency & Low & High \\
\midrule
\textbf{Primary Evaluation Metric} & mAP@0.5 & mAP@0.5:0.95 (COCO Metric) \\
\bottomrule
\multicolumn{3}{l}{\small $^{*}$\textit{Note:} The remaining 1,580 images belong to the challenge's hidden test set, for which ground truth annotations are not public.}
\end{tabular}
\end{table*}

\subsection{Real-Time Inference and Hardware-Aware Constraints}
The evaluation of real-time detectors has increasingly shifted toward hardware-aware metrics, such as latency-mAP trade-offs and energy efficiency \cite{yolov8_ultralytics, sapkota2025yolo26}. Standard GPU-based benchmarking often masks the hidden computational overhead introduced by post-processing heuristics. As documented in real-time deployment analyses \cite{lv2024rt, sapkota2025yolo26}, Non-Maximum Suppression (NMS) introduces a highly non-deterministic execution tail during inference; because its processing loops dynamically scale based on the density of candidate bounding boxes within a scene, it yields volatile frame-to-frame latency variances that disrupt strict synchronous execution streams on resource-constrained edge hardware. Furthermore, the transition from YOLOv8's Distribution Focal Loss (DFL) to YOLOv26's direct regression is strongly supported by empirical findings in model quantization \cite{gholami2022survey}. Prior deployment literature indicates that probability-distribution-based heads incur significant precision loss and inference overhead when converted to low-bitwidth integer formats (e.g., INT8) for mobile NPUs due to the non-linear Softmax operators required to resolve continuous distributions \cite{yu2025q, Shrestha2024Jul}. By adopting an NMS-free, DFL-free architecture, YOLOv26 aligns with the ``Deployment-Aware Design'' philosophy, which prioritizes the reduction of operator diversity to maximize hardware-accelerated throughput.

\subsection{Why This Comparative Analysis Matters}
The significance of comparing YOLOv8 and YOLO26 on Pascal VOC and VisDrone extends beyond incremental benchmarking. First, dataset-specific stress testing reveals architectural failure modes invisible on MS COCO: VisDrone's extremely small-object density tests the efficacy of YOLO26's STAL mechanism, while VOC's legacy taxonomy tests generalization without COCO-specific pretraining biases. Second, hardware-context sensitivity challenges the assumption that YOLO26 universally dominates: recent empirical reports indicate that at high resolutions (1280px) on server-grade GPUs (V100), YOLO26 exhibits 20\% slower inference and 37\% longer training epochs than YOLOv8, with higher memory overhead forcing reduced batch sizes. Third, as a newly released architecture, YOLO26's performance on non-COCO benchmarks remains largely unreported; this study provides the first independent evaluation on aerial small-object detection, a critical deployment domain for NMS-free architectures where deterministic latency is paramount for UAV navigation safety.
\section{Methodology}
\subsection{Overview of YOLOv8}
YOLOv8, released by Ultralytics in 2023, introduced several architectural refinements over its predecessors \cite{redmon2016you, redmon2017yolo9000, redmon2018yolov3, yolov5}. The model employs a C2f backbone (replacing YOLOv5's C3 module) that enhances gradient flow via cross-stage partial connections and a Path Aggregation Network (PANet) neck for multi-scale feature fusion. A key innovation in YOLOv8 was the transition to a decoupled detection head that separates classification and regression tasks into distinct branches. 

\textbf{Loss Function:} YOLOv8's training objective combines three components:
\begin{equation}
\mathcal{L}_{\text{YOLOv8}} = \lambda_{\text{cls}} \mathcal{L}_{\text{cls}} + \lambda_{\text{box}} \mathcal{L}_{\text{box}} + \lambda_{\text{dfl}} \mathcal{L}_{\text{DFL}}
\label{eq:yolov8-loss}
\end{equation}

where $\mathcal{L}_{\text{cls}}$ utilizes Binary Cross-Entropy (BCE) for multi-label classification.
\begin{equation}
\mathcal{L}_{\text{cls}} = -\frac{1}{N_{\text{pos}}} \sum_{i} \sum_{c} \left[ y_{i,c} \log(\hat{y}_{i,c}) + (1-y_{i,c})\log(1-\hat{y}_{i,c}) \right]
\label{eq:bce-loss}
\end{equation}

The bounding box regression loss $\mathcal{L}_{\text{box}}$ uses Complete Intersection over Union (CIoU):
\begin{equation}
\mathcal{L}_{\text{box}} = 1 - \text{IoU} + \frac{\rho^2(b, b^{\text{gt}})}{c^2} + \alpha v
\label{eq:ciou-loss}
\end{equation}

where $\rho$ denotes Euclidean distance between box centers, c  is the diagonal length of the smallest enclosing box, and v  measures aspect ratio consistency with weight α .
Distribution Focal Loss (DFL). YOLOv8 models bounding box coordinates as discrete distributions rather than direct regressions. For a coordinate y , DFL discretizes the regression range into n  bins (typically n=16 ) and computes:
\begin{equation}
\hat{y}_{\text{DFL}} = \sum_{i=0}^{n} i \cdot \text{Softmax}(w_i) = \sum_{i=0}^{n} i \cdot \frac{e^{w_i}}{\sum_{j=0}^{n} e^{w_j}}
\label{eq:dfl}
\end{equation}

where $w_i$ are learnable weights. While DFL improves localization accuracy by modeling uncertainty, the repeated Softmax operations create quantization bottlenecks on integer-arithmetic hardware.\\

\textbf{Post-Processing:} During inference, YOLOv8 generates multiple candidate boxes per object through its anchor-free, multi-scale head. NMS filters these using:
\begin{equation}
s_i = 
\begin{cases} 
s_i, & \text{if } \text{IoU}(M, b_i) < N_t \\
0, & \text{if } \text{IoU}(M, b_i) \geq N_t
\end{cases}
\label{eq:nms}
\end{equation}

where $M$ is the maximum-confidence box, $b_i$ are competing boxes, and $N_t$ is the IoU threshold (typically 0.5--0.7). This sequential dependency creates latency that scales with object density.

\subsection{Overview of YOLO26}
\begin{figure}[t]
    \centering
    \includegraphics[width=\columnwidth]{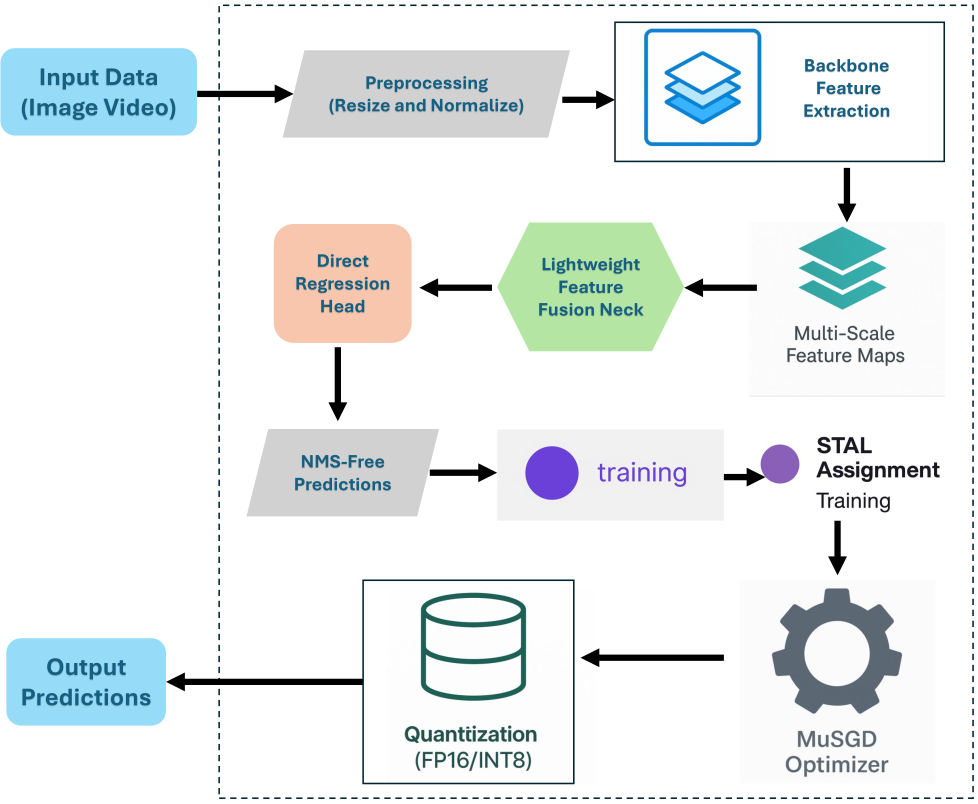}
    \caption{Architectural model of YOLO26.\cite{Sharma2026v26}}
    \label{fig:YOLO26}
\end{figure}

YOLO26 departs from the YOLOv8 paradigm through three architectural ruptures: NMS elimination, DFL removal, and dynamic training supervision \cite{chakrabarty2026YOLO26} (See \autoref{fig:YOLO26}).\\


\textbf{End-to-End NMS-Free Architecture:} YOLO26 redesigns the prediction head to enforce one-to-one label assignment during training. Instead of generating multiple candidates per object and filtering post-hoc, the model learns to emit exactly one prediction per ground-truth instance. The inference pipeline becomes a direct mapping:

\begin{equation}
\mathcal{F}_{\text{YOLO26}}: \mathbf{x} \in \mathbb{R}^{H \times W \times 3} \rightarrow \{(\hat{b}_i, \hat{c}_i)\}_{i=1}^{K}
\label{eq:e2e-mapping}
\end{equation}

where K is the number of detected objects with no post-processing stage. This eliminates Equation \ref{eq:nms} entirely, achieving deterministic latency regardless of scene complexity.\\

\textbf{Direct Regression Head (DFL-Free):} YOLO26 removes DFL in favor of direct coordinate regression:
\begin{equation}
\hat{y}_{\text{v26}} = \mathcal{F}_{\text{reg}}(\mathbf{x}) \in \mathbb{R}
\label{eq:direct-reg}
\end{equation}

\begin{equation}
\text{Head}(\mathbf{x}) = \{\mathcal{F}_{\text{cls}}(\mathbf{x}), \mathcal{F}_{\text{reg}}(\mathbf{x})\}
\label{eq:decoupled-head-v26}
\end{equation}

where Fcls predicts class probabilities, and Freg predicts box parameters directly. The decoupled structure is preserved from YOLOv8, but the regression branch bypasses distributional modeling entirely.\\

\textbf{Small-Target-Aware Label Assignment (STAL):} To address small-object vanishing, YOLO26 replaces the fixed IoU threshold with a dynamic, scale-aware assignment criterion:

\begin{equation}
\tau_{\text{dynamic}} = \tau_{\text{base}} \cdot \left(1 - \alpha \cdot e^{-\frac{\text{Area}_{\text{obj}}}{\text{Area}_{\text{img}}}}\right)
\label{eq:stal}
\end{equation}

where α  controls decay rate. For tiny objects, the exponential term approaches 1, significantly lowering τdynamic and allowing anchors with minimal physical overlap to receive positive supervision. This acts as a ``magnifying glass" for gradient signals on small targets.

Progressive Loss Balancing (ProgLoss). YOLO26 introduces time-dependent loss modulation:
\begin{equation}
\mathcal{L}_{\text{total}}(t) = \lambda_t \cdot \mathcal{L}_{\text{cls}} + (1 - \lambda_t) \cdot \mathcal{L}_{\text{box}}
\label{eq:progloss}
\end{equation}

\begin{equation}
\lambda_t = \lambda_{\text{max}} \cdot \cos\left(\frac{\pi t}{2T}\right) + \lambda_{\text{min}}
\label{eq:lambda-schedule}
\end{equation}
where $t$  is the current epoch, and T is the total training epochs. Early training (high $\lambda_t$) prioritizes semantic feature learning; late training (low $\lambda_t$) emphasizes geometric precision. This scheduling is critical for end-to-end architectures that lack the geometric guidance of anchor priors.\\

\textbf{MuSGD Optimizer:} YOLO26 employs the Muon optimizer variant (MuSGD) for stable training of lightweight backbones. Unlike standard SGD, MuSGD maintains spectral normalization constraints during gradient updates, preventing feature collapse in the CSP-Muon backbone:
\begin{equation}
\mathbf{W}_{t+1} = \mathbf{W}_t - \eta \cdot \text{SpectralNorm}(\nabla_{\mathbf{W}} \mathcal{L}_{\text{total}})
\label{eq:musgd}
\end{equation}

\autoref{tab:arch-comparison} provides a side-by-side comparison of YOLOv8 and YOLO26 across architectural components.
\begin{table*}[h!]
\centering
\caption{Architectural Comparison: YOLOv8 vs. YOLO26}
\label{tab:arch-comparison}
\resizebox{\textwidth}{!}{
\begin{tabular}{@{}lll@{}}
\toprule
\textbf{Component} & \textbf{YOLOv8} & \textbf{YOLO26} \\
\midrule
Backbone & C2f (Cross-Stage Partial with Fusion) & CSP-Muon (Edge-Optimized, Spectral Constrained) \\
Neck & PANet (Path Aggregation Network) & PANet (retained, optimized for edge export) \\
Detection Head & Decoupled (Cls + Reg + DFL branches) & Decoupled (Cls + Direct Reg, 1-to-1 assignment) \\
Anchor Strategy & Anchor-free, multi-positive assignment & Anchor-free, one-to-one assignment \\
\midrule
Classification Loss & BCE Loss (Eq. \ref{eq:bce-loss}) & BCE Loss (Eq. \ref{eq:bce-loss}) \\
Regression Loss & CIoU Loss (Eq. \ref{eq:ciou-loss}) & CIoU Loss (Eq. \ref{eq:ciou-loss}) \\
Distribution Modeling & DFL (Eq. \ref{eq:dfl}) & \textbf{Removed} (Direct Reg, Eq. \ref{eq:direct-reg}) \\
Assignment Strategy & Task Alignment Learning (TAL) & STAL (Eq. \ref{eq:stal}) \\
Dynamic Supervision & Fixed loss weights & ProgLoss (Eq. \ref{eq:progloss}, \ref{eq:lambda-schedule}) \\
Optimizer & SGD with momentum & MuSGD (Eq. \ref{eq:musgd}) \\
\midrule
Post-Processing & NMS (Eq. \ref{eq:nms}, sequential, heuristic) & \textbf{None} (native end-to-end) \\
Latency Characteristic & Variable (scales with object density) & Reduced post-processing dependency \\

Edge Quantization & DFL Softmax creates INT8 friction & Direct regression, quantization-friendly \\
\bottomrule
\end{tabular}
}
\end{table*}
\section{Experimental Setup and Benchmarking}
In this section, we discuss in detail our experimental setup and benchmark.
\subsection{Datasets and Data Preparation}
This study evaluates YOLOv8 and YOLO26 on Pascal VOC and VisDrone to assess performance under two detection settings: general object detection and small-object detection in UAV imagery. Both datasets are introduced in Section~2.2; therefore, this section focuses on their experimental use.

For VisDrone, we used the official VisDrone2019-DET train, validation, and test-dev splits, resulting in 6,471 training images, 548 validation images, and 1,610 test images. For Pascal VOC, we used the official VOC2012 training split for model training and randomly divided the official validation split into validation and test subsets using a fixed seed of 42. This produced 5,717 training images, 2,911 validation images, and 2,912 test images.

For both datasets, annotations were converted to YOLO detection format, where each object instance is represented by a class label and normalized bounding-box coordinates. During conversion, ignored regions and the ``others'' category in VisDrone were excluded. Images were resized to $640 \times 640$ and processed through the default preprocessing pipeline provided by the Ultralytics YOLO framework, which handles image loading, resizing, normalization, batching, and other framework-level operations required for YOLO-based detection. The same configuration was applied across all model variants to ensure that performance differences reflect model architecture and scale rather than differences in data preparation.

\subsection{Model Selection}
\label{model_selection}
In this study, we benchmark two YOLO model families: YOLOv8, which uses an NMS-based post-processing approach, and YOLO26, which adopts an NMS-free detection design. Each family was evaluated at five scales: nano (n), small (s), medium (m), large (l), and extra-large (x), yielding ten models in total. Matching scales across families ensures that comparisons isolate architectural differences rather than differences in model capacity.

\subsection{Training Configuration}
All models were trained using the Ultralytics YOLO framework under the same training settings to ensure a fair comparison. For each dataset, models were trained using an input image size of $640 \times 640$, a batch size of 16, and 100 training epochs. The default optimizer, learning-rate schedule, data augmentation, and loss functions provided by the Ultralytics framework were used. All models were initialized using their corresponding pretrained weights and fine-tuned on each dataset. For each training run, the best-performing checkpoint based on validation performance was selected for final evaluation.

\subsection{Experimental Environment}
All experiments were conducted with an NVIDIA RTX 6000 Ada Generation GPU with 49 GB of GPU memory. The system used CUDA version 12.9. The same hardware environment was used across all YOLOv8 and YOLO26 variants to ensure consistent training, evaluation, and latency measurements.

\subsection{Evaluation Protocol}
All models were evaluated on the validation and test splits using the built-in evaluation pipeline provided by the Ultralytics YOLO framework. Identical input sizes, inference settings, and evaluation procedures were applied across all models to ensure a consistent comparison.

The evaluation metrics were grouped into two categories: \textbf{detection accuracy} and \textbf{computational efficiency}. Detection accuracy was reported using precision, recall, F1-score, mAP@50, and mAP@50:95. Precision measures the fraction of predicted detections that are correct, while recall measures the fraction of ground-truth objects that are successfully detected. The F1-score summarizes the balance between precision and recall. The mAP@50 metric evaluates performance at an IoU threshold of 0.50, while mAP@50:95 provides a stricter assessment by averaging performance across IoU thresholds from 0.50 to 0.95.

Per-class metrics and confusion matrices were also examined to identify class-level strengths, weaknesses, and common misclassification patterns. Computational efficiency was assessed using model size, parameter count, CPU latency, GPU latency, and GFLOPs. These metrics were included because object detection models are often deployed in real-time or resource-constrained environments, where accuracy must be balanced with speed and computational cost. CPU and GPU latency were measured using the same Ultralytics evaluation/benchmarking procedure for all models.

\subsection{Pareto Frontier Analysis}
To examine the accuracy-efficiency trade-off, Pareto frontier plots were generated by comparing mAP@50:95 against GPU latency, CPU latency, model size, and parameter count. A model was considered Pareto-efficient if no other model achieved both higher accuracy and lower computational cost. This analysis is useful for comparing model families because a larger model may achieve higher accuracy, but may not be preferable if the improvement comes with substantial latency, memory, or computational cost.
\section{Experimental Results and Analysis}

We evaluate YOLOv8 and YOLO26 on the Pascal VOC and VisDrone benchmarks across five model scales: nano, small, medium, large, and extra-large. The evaluation considers detection accuracy, computational efficiency, and qualitative detection behavior. Pascal VOC is used as a general object-detection benchmark, while VisDrone is used to test performance in a more difficult aerial setting with small objects, dense scenes, and frequent occlusion. This structure allows us to compare not only which model performs better, but also how model scale, dataset difficulty, and deployment cost affect the final benchmark results.

\subsection{Overall Performance Across Model Scales}

Tables~\ref{tab:voc_detection_results} and~\ref{tab:visdrone_detection_results} present the test-set detection results for YOLOv8 and YOLO26 on Pascal VOC and VisDrone. The first clear pattern is the large performance gap between the two datasets. Both model families achieve higher scores on Pascal VOC than on VisDrone across all scales and metrics. This shows that the two benchmarks place different demands on the detectors: Pascal VOC contains more visible and less crowded object instances, while VisDrone contains many small and densely distributed targets captured from aerial viewpoints.

On Pascal VOC, YOLO26 shows a consistent advantage over YOLOv8 across model scales. The best overall result is achieved by YOLO26-x, which records the highest precision, F1-score, mAP@50, and mAP@50:95. However, the difference between YOLO26-l and YOLO26-x is very small, especially for mAP@50:95. This suggests that most of the Pascal VOC accuracy gain is already captured by the large variant, while the extra-large variant provides only a marginal additional improvement.

On VisDrone, the comparison is less separated. YOLO26 still reaches the highest overall values at the extra-large scale, but the margin over YOLOv8 is smaller than on Pascal VOC. Both model families show much lower mAP@50 and mAP@50:95 values on VisDrone, and the improvement from increasing model scale is more gradual. This indicates that the aerial small-object benchmark remains challenging for both architectures, even when model capacity is increased.

\begin{table}[t]
\centering
\caption{Test-set detection performance on Pascal VOC. Best values are shown in bold.}
\label{tab:voc_detection_results}
\footnotesize
\setlength{\tabcolsep}{4pt}
\begin{tabular}{llccccc}
\toprule
\textbf{Model} & \textbf{Scale} & \textbf{Prec.} & \textbf{Rec.} & \textbf{F1} & \textbf{mAP50} & \textbf{mAP50:95} \\
\midrule
\multirow{5}{*}{YOLOv8}
& n & 0.785 & 0.667 & 0.721 & 0.738 & 0.540 \\
& s & 0.823 & 0.689 & 0.750 & 0.762 & 0.574 \\
& m & 0.834 & 0.692 & 0.757 & 0.779 & 0.599 \\
& l & 0.818 & 0.714 & 0.762 & 0.778 & 0.602 \\
& x & 0.827 & 0.704 & 0.761 & 0.783 & 0.604 \\
\midrule
\multirow{5}{*}{YOLOv26}
& n & 0.786 & 0.688 & 0.734 & 0.750 & 0.569 \\
& s & 0.838 & 0.695 & 0.760 & 0.772 & 0.603 \\
& m & 0.835 & 0.721 & 0.774 & 0.790 & 0.619 \\
& l & 0.840 & \textbf{0.733} & 0.783 & 0.800 & 0.634 \\
& x & \textbf{0.846} & \textbf{0.733} & \textbf{0.785} & \textbf{0.801} & \textbf{0.635} \\
\bottomrule
\end{tabular}
\end{table}
\begin{table}[t]
\centering
\caption{Test-set detection performance on VisDrone. Best values are shown in bold.}
\label{tab:visdrone_detection_results}
\footnotesize
\setlength{\tabcolsep}{4pt}
\begin{tabular}{llccccc}
\toprule
\textbf{Model} & \textbf{Scale} & \textbf{Prec.} & \textbf{Rec.} & \textbf{F1} & \textbf{mAP50} & \textbf{mAP50:95} \\
\midrule
\multirow{5}{*}{YOLOv8}
& n & 0.398 & 0.304 & 0.345 & 0.268 & 0.149 \\
& s & 0.443 & 0.355 & 0.394 & 0.315 & 0.179 \\
& m & 0.498 & 0.367 & 0.423 & 0.345 & 0.200 \\
& l & 0.508 & 0.393 & 0.443 & 0.367 & 0.214 \\
& x & 0.517 & 0.395 & 0.448 & 0.368 & 0.214 \\
\midrule
\multirow{5}{*}{YOLOv26}
& n & 0.388 & 0.307 & 0.343 & 0.262 & 0.142 \\
& s & 0.454 & 0.363 & 0.403 & 0.325 & 0.182 \\
& m & 0.507 & 0.395 & 0.444 & 0.367 & 0.212 \\
& l & 0.517 & 0.398 & 0.449 & 0.372 & 0.215 \\
& x & \textbf{0.520} & \textbf{0.411} & \textbf{0.459} & \textbf{0.383} & \textbf{0.224} \\
\bottomrule
\end{tabular}
\end{table}

\subsection{Precision, Recall, and F1-Score Analysis}

The precision, recall, and F1-score results provide a closer view of the detection balance between correct predictions and missed objects. On Pascal VOC, YOLO26 generally maintains stronger precision and recall than YOLOv8, especially from the small to extra-large variants. This produces higher F1-scores for YOLO26 across most model scales, showing that its improvement is not limited to one side of the precision--recall trade-off.

For YOLOv8, performance also improves with scale, but the gains become smaller at the larger variants. The Pascal VOC results show that YOLOv8-l and YOLOv8-x are very close in F1-score, while YOLO26-l and YOLO26-x also differ only slightly. This pattern supports the broader scaling trend observed in the main results: increasing model size improves detection performance, but the benefit becomes less pronounced at the largest scales.

On VisDrone, precision, recall, and F1-score are lower for both model families. YOLO26-x gives the strongest F1-score on this benchmark, but the difference from YOLOv8-x is modest. This shows that YOLO26 improves the detection balance on VisDrone, but the improvement is smaller than the one observed on Pascal VOC. The lower scores across both models also show that VisDrone places stronger pressure on object coverage and correct localization because of its small and crowded targets.

\subsection{Localization Accuracy: mAP@50 vs. mAP@50:95}

The mAP@50 and mAP@50:95 results further clarify the difference between general detection success and stricter localization quality. On Pascal VOC, YOLO26 achieves higher mAP@50 and mAP@50:95 than YOLOv8 across all scales. The strongest results are again produced by YOLO26-x, followed closely by YOLO26-l. The small difference between these two variants shows that the large model is nearly as accurate as the extra-large model on Pascal VOC.

The mAP@50:95 metric is especially useful because it evaluates performance across stricter IoU thresholds. On Pascal VOC, YOLO26 maintains its advantage under this stricter metric, which indicates stronger bounding-box localization in addition to correct object detection. The improvement is visible at both the smaller and larger scales, although the largest gain does not always come from moving to the extra-large model.

On VisDrone, both mAP@50 and mAP@50:95 remain much lower. YOLO26-x achieves the highest VisDrone mAP values, but the margin over YOLOv8-x is limited. The lower mAP@50:95 values across both architectures show that precise localization is more difficult in aerial small-object scenes. The comparison between Pascal VOC and VisDrone therefore shows that YOLO26’s localization advantage is clearer on the general object-detection benchmark than on the dense aerial benchmark.

\subsection{Accuracy and Efficiency Trade-offs}

\begin{table}[htbp]
\centering
\caption{Computational efficiency of YOLOv8 and YOLOv26 across model scales. Latency is reported in milliseconds (ms).}
\label{tab:efficiency_results}
\scriptsize
\setlength{\tabcolsep}{3pt} 
\begin{tabular}{llccccccc}
\toprule
& & & & & \multicolumn{2}{c}{\textbf{VOC Latency}} & \multicolumn{2}{c}{\textbf{VisDrone Lat.}} \\
\cmidrule(r){6-7} \cmidrule(l){8-9}
\textbf{Model} & \textbf{Scale} & \textbf{Params} & \textbf{Size} & \textbf{GFLOPs} & \textbf{GPU} & \textbf{CPU} & \textbf{GPU} & \textbf{CPU} \\
& & \textbf{(M)} & \textbf{(MB)} & & \textbf{(ms)} & \textbf{(ms)} & \textbf{(ms)} & \textbf{(ms)} \\
\midrule
\multirow{5}{*}{YOLOv8}
& n & 3.16  & 6.25  & 8.21   & 6.65  & 21.58  & 5.52 & 16.89  \\
& s & 11.17 & 21.54 & 28.69  & 6.92  & 42.27  & 5.55 & 33.87  \\
& m & 25.90 & 49.72 & 79.13  & 7.69  & 90.01  & 6.19 & 77.71  \\
& l & 43.69 & 83.73 & 165.48 & 9.50  & 164.54 & 7.41 & 150.29 \\
& x & 68.23 & 130.55& 258.22 & 11.95 & 236.37 & 8.75 & 216.21 \\
\midrule
\multirow{5}{*}{YOLOv26}
& n & 2.57  & 5.29  & 5.81   & 7.99  & 26.67  & 7.03 & 26.75  \\
& s & 10.01 & 19.48 & 22.58  & 8.38  & 44.25  & 7.15 & 46.60  \\
& m & 21.90 & 42.21 & 74.88  & 8.55  & 103.04 & 7.33 & 102.04 \\
& l & 26.30 & 50.75 & 93.28  & 11.43 & 131.67 & 9.03 & 117.78 \\
& x & 58.99 & 113.17& 208.76 & 13.41 & 219.05 & 8.60 & 182.32 \\
\bottomrule
\end{tabular}
\end{table}


Table~\ref{tab:efficiency_results} reports the computational profile of each model, including parameter count, model size, GFLOPs, and CPU/GPU latency. These results are important because the best-performing model in accuracy is not always the most practical model for real-time deployment. Across comparable scales, YOLO26 generally uses fewer parameters, smaller model sizes, and lower GFLOPs than YOLOv8. This is especially visible at the large and extra-large scales, where YOLO26 reduces the computational footprint while still achieving strong accuracy on Pascal VOC.

\begin{figure}[h!]
    \centering
    \includegraphics[width=\columnwidth]{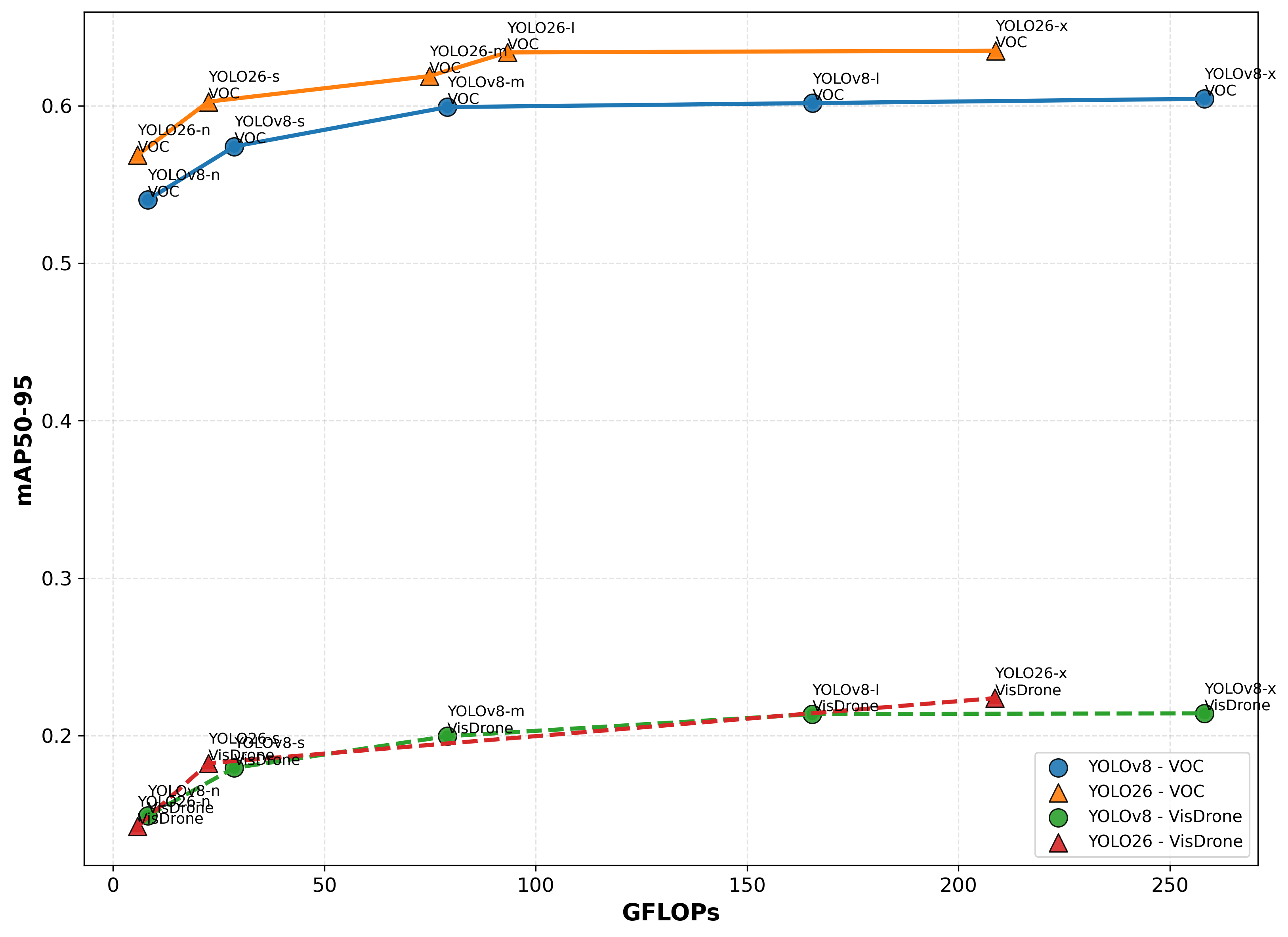}
    \caption{Accuracy-computation Pareto frontier (mAP@50:95 vs. GFLOPs) for YOLOv8 and YOLO26 on Pascal VOC and VisDrone. The curves show that accuracy generally improves with model scale, but the gain becomes smaller at the largest variants.}
    \label{fig:pareto_gflops}
\end{figure}

\begin{figure}[h!]
    \centering
    \includegraphics[width=\columnwidth]{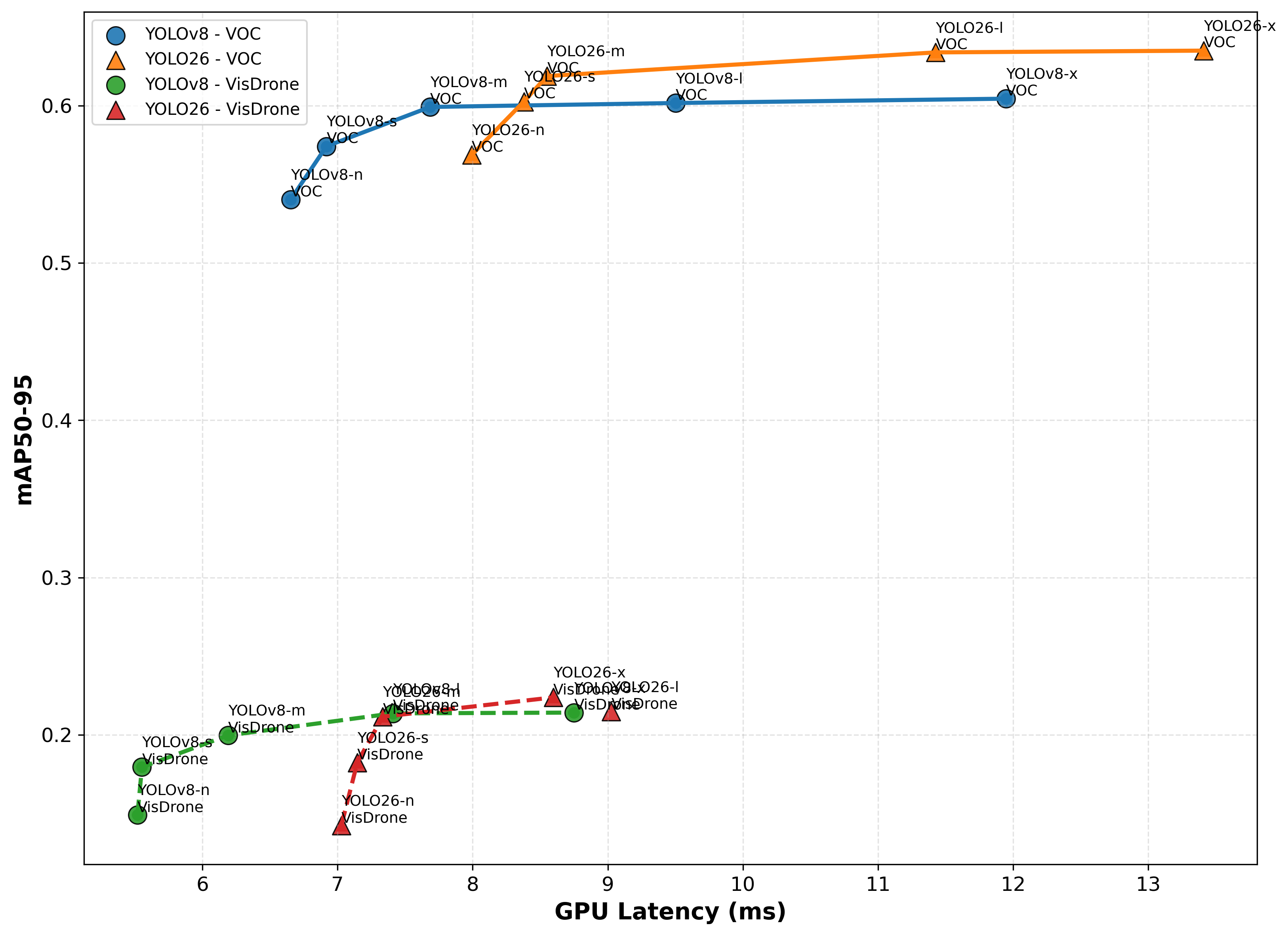}
    \caption{Accuracy-speed trade-off (mAP@50:95 vs. GPU latency) for YOLOv8 and YOLO26. The plot shows that GPU latency increases with model scale, but the increase is more moderate than the CPU latency trend.}
    \label{fig:pareto_gpu_latency}
\end{figure}

The latency results show that computational efficiency is not captured by model size or GFLOPs alone. On GPU, YOLOv8 is often faster at comparable scales, even when YOLO26 has fewer parameters or lower GFLOPs. This means that deployment performance depends not only on the number of operations, but also on how the model executes on the target hardware. On CPU, latency increases sharply for both model families as scale grows, with the largest variants carrying the highest inference cost.

Figure~\ref{fig:pareto_gflops} shows the relationship between mAP@50:95 and GFLOPs. The plot shows the expected scaling trend: larger models usually achieve higher accuracy, but the gain becomes smaller at the highest scales. This is most visible from the large to extra-large variants, where the increase in computational cost is larger than the corresponding accuracy improvement.

\begin{figure}[h!]
    \centering
    \includegraphics[width=\columnwidth]{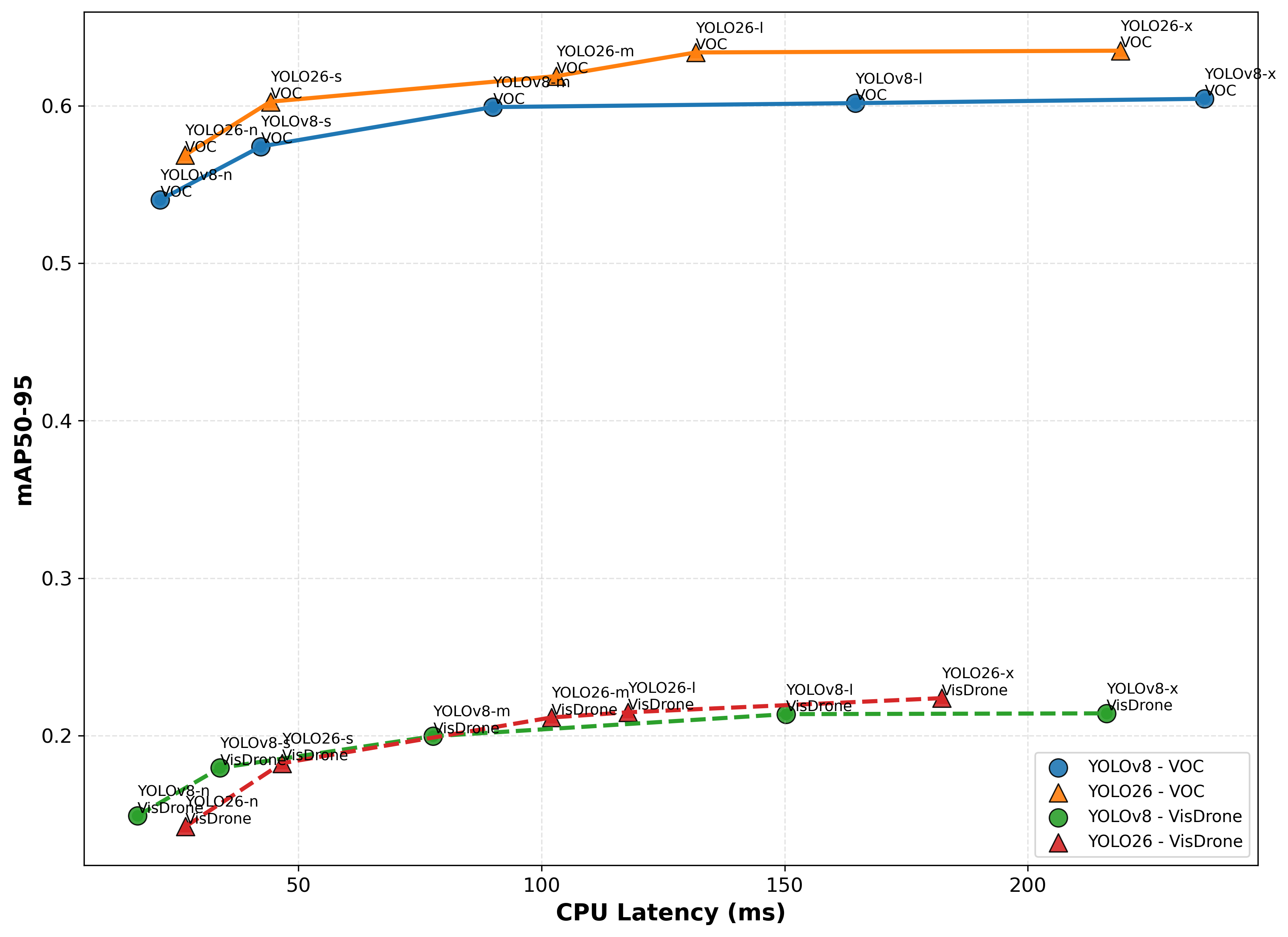}
    \caption{Accuracy-speed trade-off (mAP@50:95 vs. CPU latency) for YOLOv8 and YOLO26. The plot shows that CPU latency increases sharply with model scale, especially for large and extra-large variants.}
    \label{fig:pareto_cpu_latency}
\end{figure}

\begin{figure}[h!]
    \centering
    \includegraphics[width=\columnwidth]{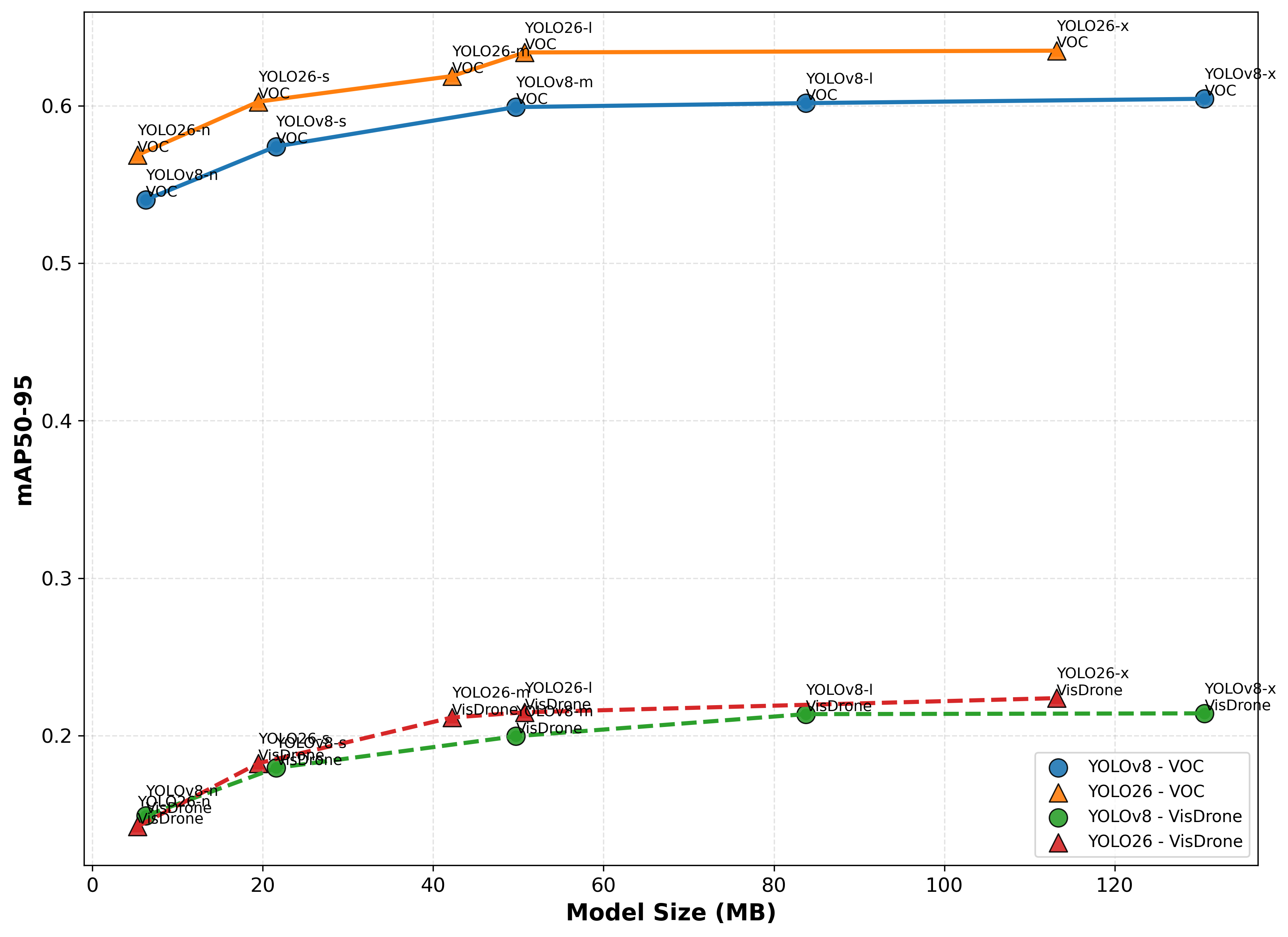}
    \caption{Accuracy-storage trade-off (mAP@50:95 vs. model size) for YOLOv8 and YOLO26. Model size increases across scales, while accuracy gains become smaller at the higher model tiers.}
    \label{fig:pareto_model_size}
\end{figure}

Figure~\ref{fig:pareto_gpu_latency} compares mAP@50:95 with GPU latency. The GPU results show a more moderate increase in latency as model scale grows. This makes the larger variants more feasible under GPU-accelerated conditions, although YOLOv8 remains competitive in speed at several scales.

Figure~\ref{fig:pareto_cpu_latency} shows the CPU latency trade-off. Compared with GPU latency, CPU latency increases more sharply as model scale grows. This is most noticeable for the large and extra-large variants, where the additional accuracy gain is accompanied by a much higher inference cost.

Figure~\ref{fig:pareto_model_size} shows the relationship between mAP@50:95 and model size. The trend is similar to the GFLOPs analysis: larger models tend to improve accuracy, but the improvement becomes smaller at the highest scales. This reinforces the pattern already shown in Table~\ref{tab:efficiency_results}, where model size and computational cost increase more quickly than accuracy at the largest variants.

\subsection{Qualitative Detection Comparison}

To complement the quantitative benchmark, we visually compare YOLOv8 and YOLO26 on representative Pascal VOC and VisDrone samples. The qualitative analysis focuses on the nano and extra-large variants. The nano models show how each architecture behaves under lightweight settings, while the extra-large models show how detection quality changes when model capacity is increased. Each example includes ground-truth boxes, predicted boxes, and overlay views to compare object coverage and localization.

\begin{figure}[h!]
    \centering
    \includegraphics[width=\columnwidth]{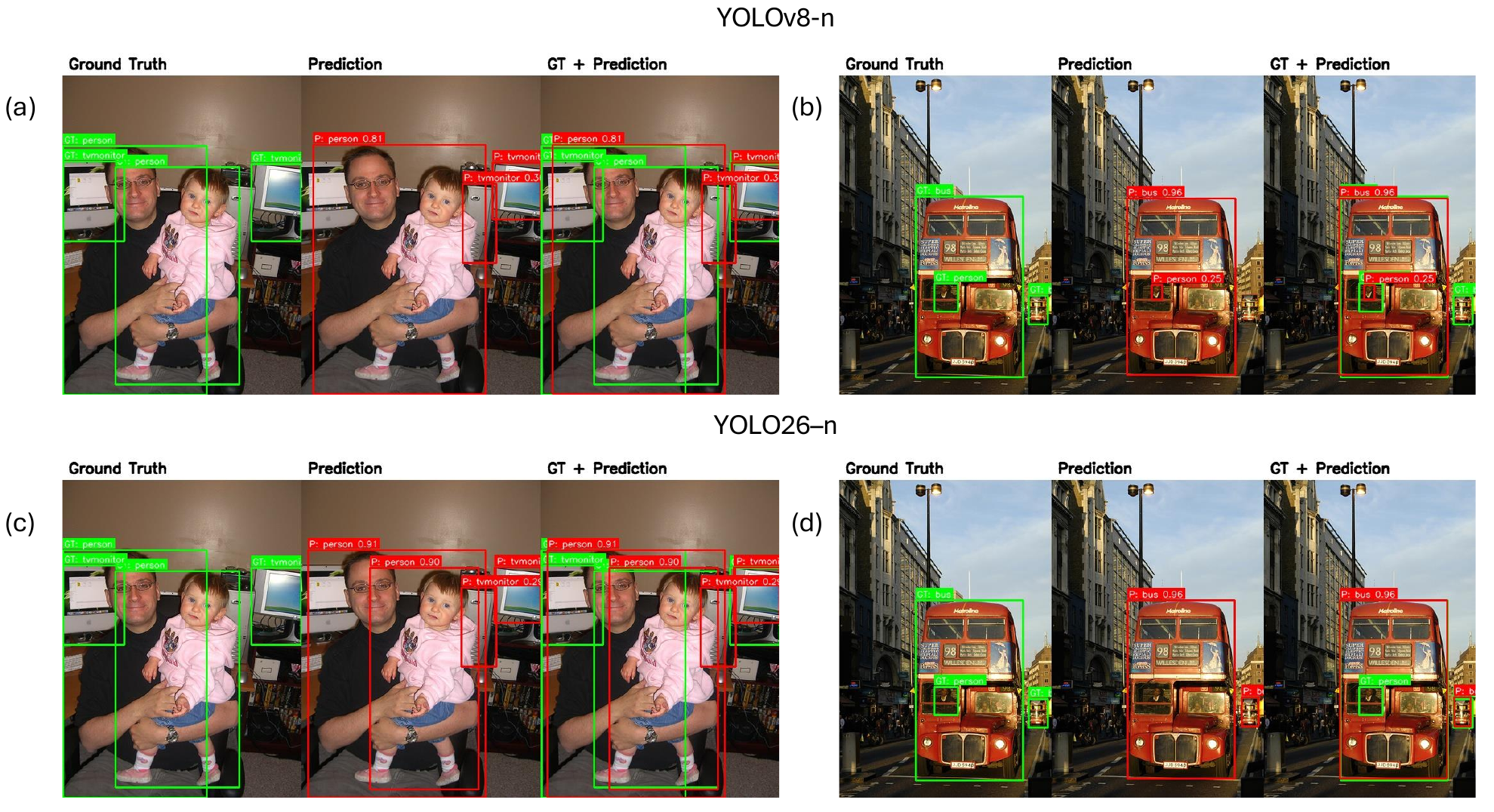}
    \caption{YOLOv8-n and YOLO26-n on Pascal VOC, showing lightweight-model behavior in standard object scenes.}
    \label{fig:qualitative_voc_nano}
\end{figure}

\begin{figure}[h!]
    \centering
    \includegraphics[width=\columnwidth]{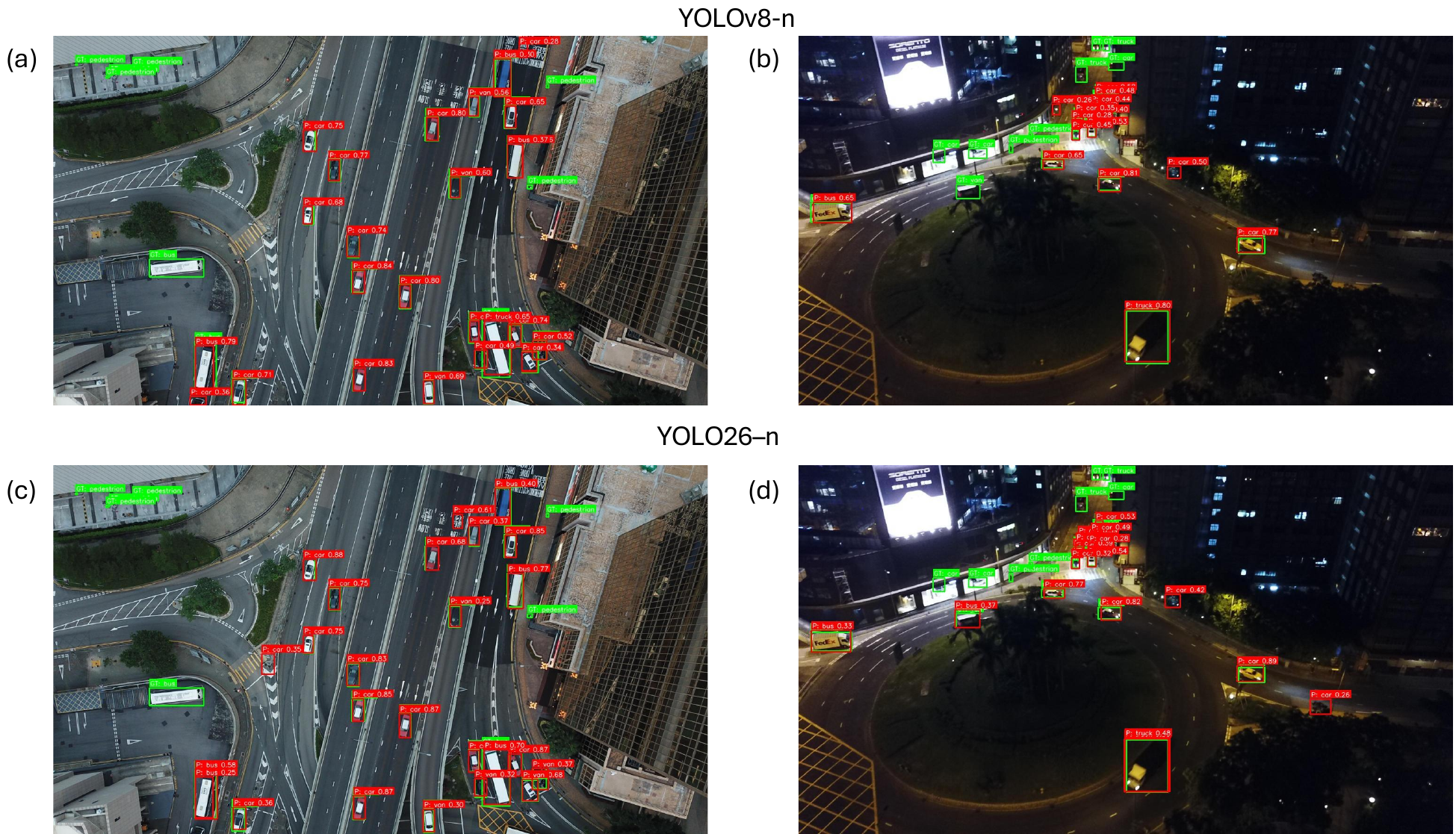}
    \caption{YOLOv8-n and YOLO26-n on VisDrone, showing lightweight detection under dense aerial small-object conditions.}
    \label{fig:qualitative_VisDrone_n}
\end{figure}

\paragraph{Nano Variants}
Figure~\ref{fig:qualitative_voc_nano} compares the nano variants on Pascal VOC. In the selected examples, both models detect the main foreground objects, but they differ in smaller-object coverage and bounding-box alignment. YOLOv8-n captures some secondary objects that YOLO26-n misses or predicts with lower confidence, while both models remain limited by the reduced capacity of the nano scale.

Figure~\ref{fig:qualitative_VisDrone_n} shows the nano variants on VisDrone. Compared with Pascal VOC, the aerial scenes contain smaller and more crowded targets, making missed detections and localization errors more visible. This visual pattern is consistent with the lower VisDrone scores reported in Table~\ref{tab:visdrone_detection_results}.

\begin{figure}[h!]
    \centering
    \includegraphics[width=\columnwidth]{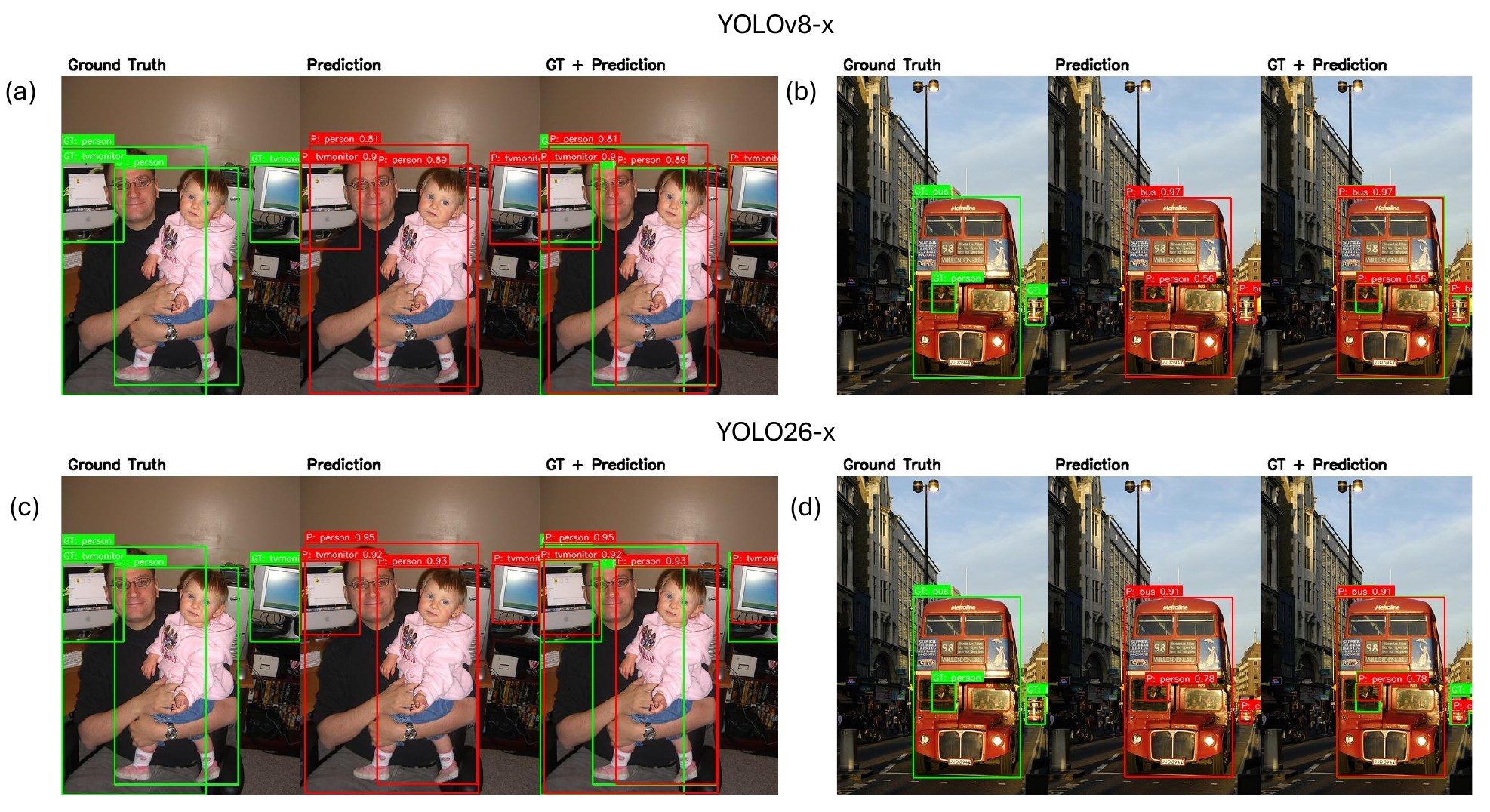}
    \caption{YOLOv8-x and YOLO26-x on Pascal VOC, showing high-capacity detection in standard object scenes.}
    \label{fig:qualitative_voc_x}
\end{figure}

\begin{figure}[hpbt!]
    \centering
    \includegraphics[width=\columnwidth]{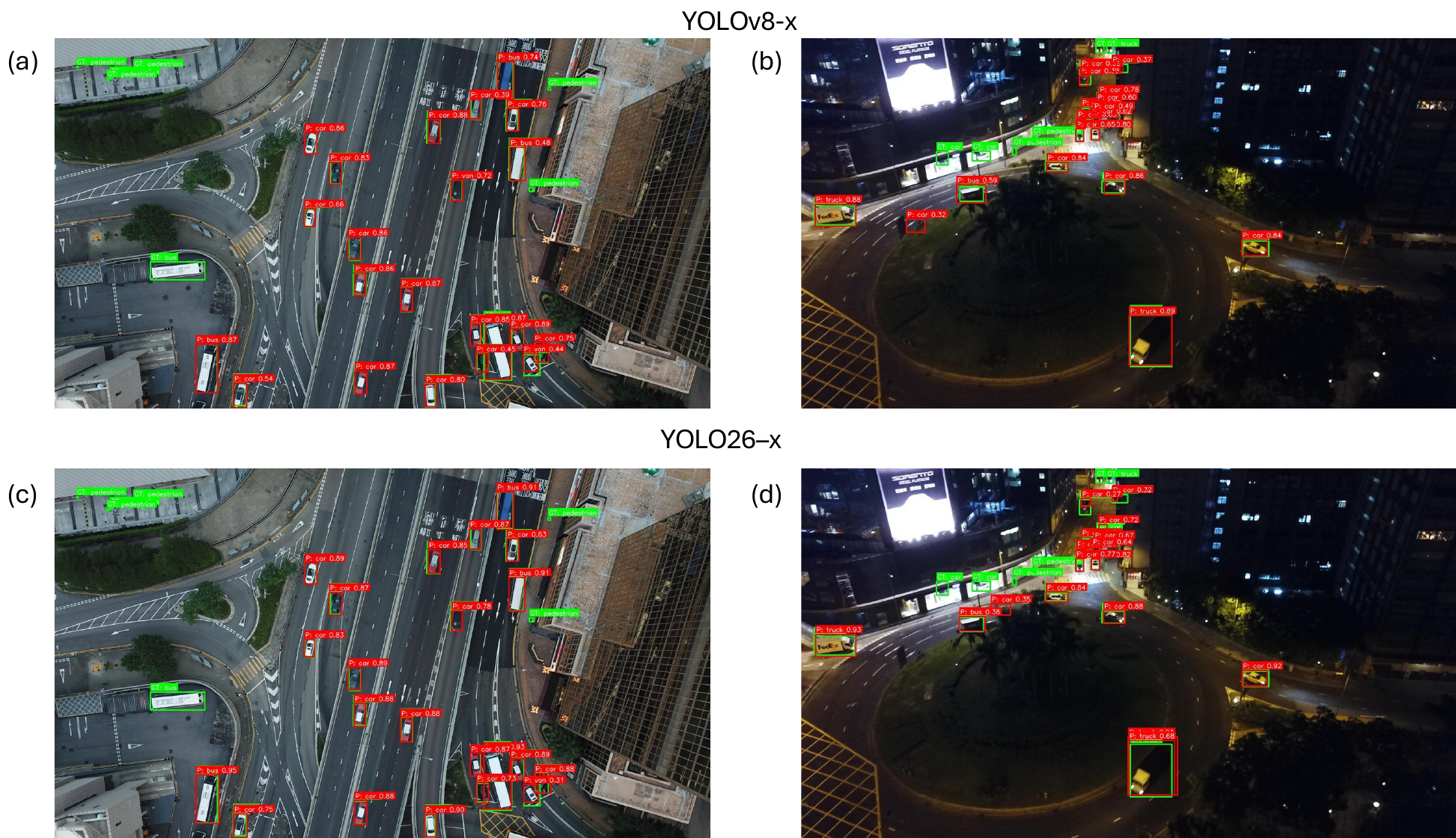}
    \caption{YOLOv8-x and YOLO26-x on VisDrone, showing high-capacity detection in crowded aerial scenes.}
    \label{fig:qualitative_visdrone_x}
\end{figure}

\paragraph{Extra-Large Variants}
Figure~\ref{fig:qualitative_voc_x} compares the extra-large variants on Pascal VOC. Relative to the nano examples, the larger models show more complete object coverage and closer alignment between predicted and ground-truth boxes. This matches the quantitative trend in Table~\ref{tab:voc_detection_results}, where the large and extra-large variants achieve the strongest mAP@50:95 values.

Figure~\ref{fig:qualitative_visdrone_x} compares the extra-large variants on VisDrone. Although the larger models improve object coverage compared with the nano variants, dense target placement and small object size still make aerial detection more challenging than Pascal VOC. 

This trend is also visible across the small, medium, and large variants of both YOLOv8 and YOLO26. The qualitative results show that detection coverage and localization generally improve as model capacity increases from nano to extra-large. On Pascal VOC, the larger variants improve the detection of both nearby and distant objects. On VisDrone, however, both model families continue to miss many small targets across variants, highlighting the difficulty of detecting small, distant objects in aerial imagery.

\section{Conclusion}
This paper benchmarked YOLOv8 and YOLO26 across Pascal VOC and VisDrone using five model scales and multiple accuracy-efficiency metrics. The results show that YOLO26 achieves stronger detection performance on Pascal VOC across most scales, with the extra-large variant producing the highest overall accuracy. However, the gain from YOLO26-l to YOLO26-x is small, indicating diminishing returns at the largest scale.

On VisDrone, both YOLOv8 and YOLO26 show substantially lower performance, and the gap between the two model families is narrower. This indicates that dense aerial small-object detection remains difficult for both architectures. The efficiency analysis further shows that YOLO26 generally reduces parameter count, model size, and GFLOPs, but latency depends on the hardware setting, with YOLOv8 remaining competitive on GPU.

Overall, the results suggest that YOLO26 is a strong alternative to YOLOv8, particularly when accuracy and model compactness are important. However, the best detector depends on the dataset, object scale, and deployment constraints rather than architecture alone.

\balance
\small
\bibliographystyle{ieeenat_fullname}
\bibliography{references}

\end{document}